\documentclass[runningheads]{llncs}
\usepackage[T1]{fontenc}
\usepackage{graphicx}
\usepackage{amsmath}
\usepackage{amssymb}
\usepackage{xcolor}
\usepackage{hyperref}
\usepackage{booktabs}
\usepackage{multirow}
\usepackage{enumitem}
\usepackage{tabularx}
\usepackage{listings}
\usepackage{ragged2e}
\usepackage{makecell}
\usepackage[nameinlink,capitalize]{cleveref}
\usepackage{tcolorbox}
\tcbuselibrary{breakable}
\usepackage{chngcntr}

\newcolumntype{Y}{>{\ttfamily\RaggedRight\arraybackslash}p{0.38\linewidth}}
\usepackage[scaled=0.92]{beramono}
\hypersetup{
    colorlinks=true,
    linkcolor=blue,
    citecolor=blue,
    urlcolor=blue
}

\newenvironment{LogicBox}[1]{%
  \begin{tcolorbox}[breakable, colback=gray!3, colframe=gray!60, boxrule=0.6pt,
                    title=\textbf{#1}, left=6pt, right=6pt, top=4pt, bottom=4pt]%
}{\end{tcolorbox}}

\newcounter{card}
\renewcommand{\thecard}{\arabic{card}}
\crefname{card}{Card}{Cards}
\Crefname{card}{Card}{Cards}

\begin{document}

\title{From Natural Language to Executable Narsese: A Neuro-Symbolic Benchmark and Pipeline for Reasoning with NARS}

\author{Mina Gabriel\inst{1} \and Pei Wang\inst{1}}
\authorrunning{M. Gabriel et al.}

\institute{Temple University, Philadelphia, PA 19122, USA\\
\email{\{mina.gabriel,pei.wang\}@temple.edu}}

\maketitle

\newcommand{\code}[1]{\texttt{#1}}

\begin{abstract}
Large language models (LLMs) are highly capable at language generation, but they remain unreliable when reasoning requires explicit symbolic structure, multi-step inference, and interpretable uncertainty. This paper presents a neuro-symbolic framework for translating natural-language reasoning problems into executable formal representations using first-order logic (FOL) and Narsese, the language of the Non-Axiomatic Reasoning System (NARS). To support this direction, we introduce \textit{NARS-Reasoning-v0.1}, a benchmark of natural-language reasoning problems paired with FOL forms, executable Narsese programs, and three gold labels: \textsc{True}, \textsc{False}, and \textsc{Uncertain}. We develop a deterministic compilation pipeline from FOL to executable Narsese and validate retained examples through runtime execution in OpenNARS for Applications (ONA), ensuring that the symbolic targets are not only syntactically well formed but also behaviorally aligned with the intended answer. We further present Language-Structured Perception (LSP), a formulation in which an LLM is trained to produce reasoning-relevant symbolic structure rather than only a final verbal response. As an initial proof of concept, we also train and release a Phi-2 LoRA adapter on \textit{NARS-Reasoning-v0.1} for three-label reasoning classification, showing that the benchmark can support supervised adaptation in addition to executable evaluation. Overall, the paper positions executable symbolic generation and execution-based validation as a practical path toward more reliable neuro-symbolic reasoning systems.

\end{abstract}

\section{Introduction}

Large language models (LLMs) have shown impressive performance in natural language processing, especially in few-shot learning and instruction-following settings. However, strong language generation should not be confused with reliable reasoning. When a task requires explicit logical structure, controlled composition, or traceable inference, neural models often fail in familiar ways: they may produce answers that sound plausible but cannot be verified, break down under distribution shift, or respond with high confidence even when the underlying logical support is missing \cite{yang2023logicllama,pan2023logiclm}.

These weaknesses are especially visible in tasks based on first-order logic (FOL). Text-only answers are often insufficient because they cannot be directly executed, audited, or checked for semantic equivalence. As a result, a system may produce a convincing answer even when its underlying representation is incorrect \cite{gabriel2025semanticgap}.

A natural response is to combine LLMs with symbolic systems. Recent work has explored LLMs as translators into logical form, LLMs coupled with theorem provers, and hybrid memory-reasoning systems that use formal representations to improve faithfulness and persistence \cite{yang2023logicllama,pan2023logiclm,isaev2025narsgpt}. We build on this direction, but focus on a different target formalism: \emph{Narsese}, the language of the Non-Axiomatic Reasoning System (NARS). Unlike classical theorem provers that assume sufficient knowledge and resources, NARS is designed for reasoning under bounded resources and incomplete knowledge, which makes it an attractive substrate for broader cognitive architectures \cite{wang2013nal}.

The central idea of this paper is that natural-language reasoning should be evaluated through an \emph{executable} neuro-symbolic pipeline. In our setting, a system is asked to read a natural-language context and query, construct a formal representation, compile it into executable Narsese, and then allow a reasoning engine to produce the final decision. This shifts the task away from verbal answer generation alone and toward the construction of symbolic structure that another system can actually use.

To make this concrete, we introduce \textit{NARS-Reasoning-v0.1}, a benchmark of 1{,}000 natural-language reasoning instances derived from ProverQA-style problems, balanced across three difficulty levels and validated by execution in OpenNARS for Applications (ONA) \cite{hammer2020ona,gabriel2025narsreasoning}. Each instance pairs natural-language reasoning content with FOL forms, executable Narsese, and one of three gold labels: \textsc{True}, \textsc{False}, or \textsc{Uncertain}. We also describe a deterministic compiler from FOL to executable Narsese and define a trainable extension, \emph{Language-Structured Perception (LSP)}, in which a model is trained to produce reasoning-relevant symbolic structure rather than only a free-form answer.

Beyond the benchmark itself, we show that the resource also supports supervised model development. As an initial proof of concept, we trained and released a Phi-2 LoRA adapter on \textit{NARS-Reasoning-v0.1} for three-label reasoning classification. This released model does not yet generate full executable symbolic programs, but it demonstrates that the benchmark is already usable as a practical training resource in addition to an executable evaluation benchmark \cite{gabriel2025phi2adapter}.

The contributions of this paper are as follows:
\begin{itemize}[leftmargin=1.5em]
    \item We introduce a difficulty-balanced benchmark of natural-language reasoning instances paired with FOL forms, executable Narsese programs, and three gold labels.
    \item We describe a deterministic FOL-to-Narsese compiler that supports a useful subset of atomic predicates, negation, conjunction, disjunction, implication, XOR, and quantified rules.
    \item We define an execution-based validation procedure in ONA that filters symbolic programs by agreement with the intended answer under a fixed runtime protocol.
    \item We formalize a neuro-symbolic learning setting in which symbolic executability becomes part of the supervision signal.
    \item We show that the benchmark can also support supervised adaptation in practice by training and releasing a Phi-2 LoRA baseline on the dataset.
\end{itemize}

Rather than treating symbolic structure as an optional interpretability aid, we treat it as the object that must be produced correctly. In this view, the benchmark tests whether a system can go from language to a representation that supports actual reasoning, not simply whether it can guess the right answer.

\section{Background and Related Work}

\subsection{Natural Language to Formal Logic}

Translating natural language into formal logic has a long history in formal semantics and computational linguistics. Classical approaches relied on hand-built grammars, semantic templates, and carefully engineered parsers. These systems were precise, but they were costly to build and difficult to scale to open-domain text \cite{zettlemoyer2005learning,bos2005recognising}.

Neural semantic parsing reduced this manual burden by learning mappings from text to structured meaning representations. More recently, LLMs have significantly improved zero-shot and few-shot performance on semantic parsing and code-like generation tasks, including translation from natural language into FOL \cite{yang2023logicllama}. LogicLLaMA is particularly relevant because it demonstrates that a relatively small fine-tuned model can correct or directly generate FOL expressions competitively, especially when training is coupled with logical validation and correction signals \cite{yang2023logicllama}.

That line of work motivates a key design choice in our paper: natural language should not be judged only by surface-form similarity to a reference logical expression. It should also be judged by whether the generated representation is \emph{executable} and leads to the correct reasoning outcome.

\subsection{LLMs with Symbolic Reasoners}

A growing line of research combines LLMs with symbolic reasoning systems to improve logical reliability. For example, Logic-LM uses a symbolic solver as an external verifier: the LLM proposes candidate reasoning structures, and a formal system checks whether they are valid \cite{pan2023logiclm}. ProverGen and ProverQA extend this idea at the dataset level by combining the linguistic flexibility of LLMs with symbolic provers that enforce logical correctness in the generated benchmark \cite{qi2025provergen}.

Our work follows this general direction, but differs in both representation and execution. Rather than stopping at FOL generation or theorem-prover evaluation, we compile supported FOL into Narsese and execute it in ONA. This makes the framework relevant not only to theorem-proving benchmarks, but also to reasoning systems, cognitive architectures, and hybrid AGI settings.

\subsection{NARS and NARS-GPT}

NARS is a reasoning framework designed for operation under insufficient knowledge and resources. It emphasizes non-axiomatic reasoning, incremental belief revision, and explicit treatment of uncertainty \cite{wang2013nal}. ONA is an implementation of NARS optimized for real-time applications and practical control \cite{hammer2020ona}.

NARS-GPT showed that GPT-style language models can be integrated with NARS to support natural-language interaction, persistent knowledge, and runtime reasoning, including symbol grounding and long-term memory updates \cite{isaev2025narsgpt}. Our work differs from NARS-GPT in priority. NARS-GPT focuses on an integrated natural-language interaction system. By contrast, we focus on benchmark construction and executable program generation: the central task here is to generate symbolic structure from natural language and evaluate it through execution.

\section{Problem Formulation}

We study the following task. Given a natural-language context $x = (c, q)$, where $c$ is a set of premises expressed in English and $q$ is a claim or query, the system must predict a label
\[
y \in \{\textsc{True}, \textsc{False}, \textsc{Uncertain}\}.
\]

However, unlike standard text classification, the desired output is not only a label. The system is expected to construct an intermediate symbolic representation that can be executed. In the most general form considered in this paper, the pipeline is
\[
(c,q) \xrightarrow{\text{NL}\to\text{FOL}} z_{\mathrm{FOL}} \xrightarrow{\text{FOL}\to\text{Narsese}} z_{\mathrm{Nar}} \xrightarrow{\text{ONA}} \hat{y}.
\]

This framing turns logical reasoning into a structured generation problem with executable semantics. A model succeeds only if the generated symbolic object preserves enough structure for the external reasoner to reach the correct outcome.

\subsection{Language-Structured Perception}

We use the term \emph{Language-Structured Perception (LSP)} for a model's ability to transform natural-language inputs into an executable reasoning structure. In LSP, the model is not rewarded solely for producing fluent or locally plausible text. Instead, it is encouraged to organize linguistic content into symbolic components such as facts, rules, queries, and negated statements that support downstream execution.

Formally, given a dataset
\[
\mathcal{D}=\{(x_i,P_i,G_i,y_i,z^{*}_{\mathrm{FOL},i})\}_{i=1}^{N},
\]
where $x_i$ denotes the natural-language input, $P_i$ the latent or explicit premise set, $G_i$ the goal statement, $y_i$ the gold label, and $z^{*}_{\mathrm{FOL},i}$ the gold FOL form, the model learns a mapping
\[
\pi_{\theta}: x_i \mapsto z_{\mathrm{FOL},i} \mapsto z_{\mathrm{Nar},i} \mapsto \hat{y}_i.
\]

The important shift is conceptual: the reasoning target is an executable structure, not entirely a textual answer.

\section{The \textit{NARS-Reasoning-v0.1} Benchmark}

\subsection{Source and Motivation}

\textit{NARS-Reasoning-v0.1} is derived from the ProverQA family of first-order logic (FOL) reasoning problems introduced through the ProverGen framework \cite{qi2025provergen}. ProverGen combines LLM-based generation with symbolic verification to produce reasoning instances that are both linguistically varied and logically coherent. This makes it a strong starting point for our setting, since the source problems already contain structured premises, explicit queries, and three-way reasoning labels.

Our goal is not to replace ProverQA, but to transform a subset of its reasoning style into a benchmark that supports executable neuro-symbolic evaluation. Each retained instance is paired not only with natural-language content and a gold label, but also with an executable Narsese program that can be run in ONA. This changes the task from static logical evaluation to runtime validation under a concrete reasoning engine. In other words, the benchmark is designed to test not only whether the answer is correct, but also whether the symbolic representation is usable by the target system.

\subsection{Dataset Composition}

\textit{NARS-Reasoning-v0.1} contains 1{,}000 instances. Each instance includes:
\begin{itemize}[leftmargin=1.5em]
    \item a natural-language context consisting of facts and rules,
    \item a natural-language claim or question,
    \item a gold label in \{\textsc{True}, \textsc{False}, \textsc{Uncertain}\},
    \item an FOL representation used for compilation, and
    \item an executable Narsese program validated in ONA.
\end{itemize}

The dataset is balanced across three reasoning difficulty levels defined by the number of inference steps required to reach the correct conclusion:
\begin{itemize}[leftmargin=1.5em]
    \item \textbf{Easy}: 1--2 reasoning steps,
    \item \textbf{Medium}: 3--5 reasoning steps,
    \item \textbf{Hard}: 6--9 reasoning steps.
\end{itemize}

The full benchmark contains 400 easy, 300 medium, and 300 hard instances. The default split uses 800 training instances and 200 test instances, with the test set balanced as 100 easy, 50 medium, and 50 hard examples.

\begin{table}[h]
\centering
\caption{Difficulty-balanced split of \textit{NARS-Reasoning-v0.1}.}
\label{tab:splits}
\begin{tabular}{lcccc}
\toprule
\textbf{Split} & \textbf{Easy} & \textbf{Medium} & \textbf{Hard} & \textbf{Total}\\
\midrule
Train & 300 & 250 & 250 & 800\\
Test & 100 & 50 & 50 & 200\\
\midrule
Total & 400 & 300 & 300 & 1000\\
\bottomrule
\end{tabular}
\end{table}

This balance is important for two reasons. First, it prevents performance from being dominated by short and shallow cases. Second, it makes it easier to study how systems degrade as the required reasoning depth increases. A model that performs well only on easy instances should not be interpreted as a model that preserves symbolic structure through deeper inference chains.

\subsection{Label Space}

We use the same three labels as the original reasoning setup:
\begin{itemize}[leftmargin=1.5em]
    \item \textsc{True}: the claim is supported by the premises,
    \item \textsc{False}: the negation of the claim is supported, or the claim is contradicted,
    \item \textsc{Uncertain}: the available information is insufficient to derive either side.
\end{itemize}

This three-label setup is especially important for neuro-symbolic evaluation. In ordinary binary settings, a system is forced to guess even when the available information is incomplete. That can hide an important distinction between being wrong and not having enough evidence. By explicitly including \textsc{Uncertain}, the benchmark separates contradiction from underspecification and makes it possible to evaluate whether a model preserves that distinction throughout the symbolic pipeline.

More generally, this label space matches the goals of executable reasoning more closely than a binary formulation. A symbolic system should not only derive supported conclusions, but also refrain from making unsupported ones. For that reason, \textsc{Uncertain} is not a fallback category; it is a necessary part of the reasoning task.

\section{Compiling FOL into Executable Narsese}

\subsection{Compiler Overview}

A central component of the benchmark is a deterministic compiler that converts first-order logic (FOL) expressions into executable Narsese. The compiler first tokenizes each FOL expression, builds an abstract syntax tree (AST), and then recursively maps the supported operators into Narsese forms. In the current benchmark, this supported subset includes atomic predicates, negation, conjunction, disjunction, implication, XOR, and the quantified forms needed by the selected examples.

Before translation, the compiler also applies a small amount of normalization. It removes prefixes such as \texttt{fact7:} and \texttt{rule5:}, strips lightweight formatting artifacts such as bullets or quotation marks, and removes trailing punctuation. This makes the conversion pipeline more robust to lightly formatted inputs.

\subsection{Core Mapping Patterns}

Table~\ref{tab:fol_to_narsese_rules} summarizes the main FOL-to-Narsese conversion patterns used in the benchmark. The mapping is intentionally mechanical. We do not claim full equivalence between unrestricted FOL and unrestricted Narsese. Instead, we define a supported subset that can be translated consistently and executed in ONA.

\begin{table}[h]
\caption{Core FOL-to-Narsese conversion patterns used in the benchmark.}
\label{tab:fol_to_narsese_rules}
\centering
\setlength{\tabcolsep}{4pt}
\renewcommand{\arraystretch}{1.08}
\footnotesize
\begin{tabularx}{\linewidth}{@{}>{\RaggedRight\arraybackslash}p{0.28\linewidth}
                                  >{\RaggedRight\arraybackslash}p{0.38\linewidth}
                                  >{\RaggedRight\arraybackslash}X@{}}
\toprule
\textbf{FOL Expression} & \textbf{Narsese Conversion} & \textbf{Notes} \\
\midrule
\multicolumn{3}{@{}l}{\textit{Atomic Statements}}\\
\addlinespace[2pt]
Atom \(p(a)\) (unary const)  & \code{<\{a\} $-->$ p>.} & Constant represented as an individual term. \\[3pt]
Atom \(p(x)\) (unary var)    & \code{<\$1 $-->$ p>.} & Variable represented as \texttt{\$1}, \texttt{\$2}, \dots \\[3pt]
Atom \(r(a,b)\) (binary)     & \code{<(\{a\} * \{b\}) $-->$ r>.} & Product term \texttt{*} forms tuples. \\[3pt]
Negation \(\lnot \phi\) (atomic) & \code{($--$ <...>).} & Statement negation uses the prefix \texttt{$--$}. \\[6pt]
\midrule
\multicolumn{3}{@{}l}{\textit{Implication and Composition}}\\
\addlinespace[2pt]
\(A \to B\) & \code{<A ==> B>.} & Default implication form. \\[4pt]
\((A \land B) \to C\) & \code{<(A \& B) ==> C>.} & Conjunctive antecedent. \\[4pt]
\((A \lor B) \to C\)  & \code{<A ==> C>.}, \code{<B ==> C>.} & Disjunctive antecedent is split into two rules. \\[4pt]
\(A \to (B \land C)\) & \code{<A ==> B>.}, \code{<A ==> C>.} & Conjunctive consequent is split into two rules. \\[4pt]
\(A \to (B \lor C)\)  & \code{<A ==> B>.}, \code{<A ==> C>.} & This conversion strengthens the original FOL form. \\[4pt]
\((A \oplus B) \to C\) & \code{<A ==> ($--$ B)>.}, \code{<B ==> ($--$ A)>.}, \code{<A ==> C>.}, \code{<B ==> C>.} & XOR expanded into exclusivity plus consequence rules. \\[6pt]
\midrule
\multicolumn{3}{@{}l}{\textit{Questions}}\\
\addlinespace[2pt]
Ask conclusion \(\phi\) & \code{<...>?} & Append \texttt{?} to the compiled statement. \\[2pt]
\bottomrule
\end{tabularx}
\end{table}

\subsection{Handling Composite Conclusions}

A practical issue arises when the target conclusion is itself composite. Since a Narsese question must be expressed as a statement-level query, the compiler first attempts to translate the full conclusion directly. If that is not possible, it falls back to the first atomic or negated atomic subformula found in the AST and uses that as the query. This fallback keeps the example executable while preserving a direct connection to the original target conclusion.

\subsection{Illustrative Example}

To make the compilation process more concrete, \Cref{card:multi-fact} presents a worked example that follows a single reasoning instance across three aligned representations: the original natural-language premises, the corresponding first-order logic (FOL) statements, and the executable Narsese program. The purpose of this example is to show how a short multi-fact reasoning problem is transformed into a symbolic form that can be executed by the reasoner. The card is organized in a step-by-step way: it first presents the premises in natural language, FOL, and Narsese, then shows the target conclusion in the same three forms, and finally gives the resulting label.

In this example, the query asks whether Jasiah is not innovative. The premises provide information about creativity, artistic behavior, and love of drawings, but they do not provide any rule or fact connecting these statements to innovativeness. As a result, neither the statement nor its negation can be derived from the available information. The correct label is therefore \textsc{Uncertain}. This example highlights the core goal of the benchmark: natural-language reasoning problems are translated into executable symbolic structure, and the final decision is determined by the behavior of the reasoner over that structure.

This example shows how the benchmark connects language, logic, and executable reasoning in a single pipeline. The natural-language input is first formalized in FOL, then compiled into Narsese, and finally evaluated by the reasoner to produce the final label.

\refstepcounter{card}
\phantomsection
\label[card]{card:multi-fact}

\begin{LogicBox}{\textbf{Card~\thecard.} Multi-Fact Natural Language to FOL and Narsese}
\small

\definecolor{nlbg}{HTML}{E8F1FA}
\definecolor{nlfg}{HTML}{1B4F72}
\definecolor{folbg}{HTML}{EAF7EE}
\definecolor{folfg}{HTML}{1D6F42}
\definecolor{nabg}{HTML}{EFEAF7}
\definecolor{nafg}{HTML}{4A3B6E}
\newcommand{\tagNL}{\colorbox{nlbg}{\strut\textcolor{nlfg}{\textsf{\footnotesize\;NL\;}}}}
\newcommand{\tagFOL}{\colorbox{folbg}{\strut\textcolor{folfg}{\textsf{\footnotesize\;FOL\;}}}}
\newcommand{\tagNAR}{\colorbox{nabg}{\strut\textcolor{nafg}{\textsf{\footnotesize\;Narsese\;}}}}

\noindent\textbf{Premises}\par\vspace{4pt}
\begingroup\setlength{\parindent}{0pt}

\tagNL\par\vspace{2pt}
Jasiah values creativity. Anyone who loves drawings and values creativity is artistic.
Jones loves drawings. Jasiah loves drawings.\par\vspace{1.2\baselineskip}

\tagFOL\par\vspace{2pt}
\(
\begin{aligned}
&\text{fact1:}\ \ \mathrm{values\_creativity}(\mathrm{Jasiah})\\
&\text{rule1:}\ \ \forall x\big(\mathrm{loves\_drawings}(x)\land \mathrm{values\_creativity}(x)\ \to\ \mathrm{artistic}(x)\big)\\
&\text{fact2:}\ \ \mathrm{loves\_drawings}(\mathrm{Jones})\\
&\text{fact3:}\ \ \mathrm{loves\_drawings}(\mathrm{Jasiah})
\end{aligned}
\)\par\vspace{1.2\baselineskip}

\tagNAR\par\vspace{2pt}
\texttt{$<${Jasiah} $-->$ values\_creativity$>$.}\\
\texttt{$<<$(\$1 $-->$ loves\_drawings) $\&\&$ (\$1 $-->$ values\_creativity)$>$ $==>$ $<$\$1 $-->$ artistic$>$$>>$.}\\
\texttt{$<${Jones} $-->$ loves\_drawings$>$.}\\
\texttt{$<${Jasiah} $-->$ loves\_drawings$>$.}\\
\texttt{($--$ $<${Jasiah} $-->$ innovative$>$)?.}
\par
\endgroup

\vspace{6pt}
\noindent\rule{\linewidth}{0.5pt}\par\vspace{6pt}

\noindent\textbf{Conclusion}\par\vspace{4pt}
\begin{tabularx}{\linewidth}{@{}p{0.16\linewidth} X@{}}
\tagNL  & Jasiah is not innovative. \\[6pt]
\tagFOL & \(\lnot innovative(\mathrm{Jasiah})\) \\[6pt]
\tagNAR & \texttt{($--$ $<${Jasiah} $-->$ innovative$>$)?} \\
\end{tabularx}

\vspace{6pt}
\noindent\rule{\linewidth}{0.5pt}\par\vspace{6pt}

\noindent\textbf{Answer}\par\vspace{2pt}
\begin{center}
{\large\textbf{Uncertain}}
\end{center}
\vspace{-4pt}

\normalsize
\end{LogicBox}

\section{Execution-Based Validation in ONA}

The Non-Axiomatic Reasoning System (NARS) is a reasoning framework designed for settings in which knowledge is incomplete and computational resources are limited. Instead of assuming a fixed set of sufficient axioms, NARS treats reasoning as an ongoing process of drawing, revising, and prioritizing conclusions under bounded conditions. Its formal language, Narsese, represents statements, questions, and goals in an explicit symbolic form, which makes it suitable for neuro-symbolic pipelines in which natural-language inputs are translated into executable programs rather than treated only as free-form text.

In this work, execution is carried out in OpenNARS for Applications (ONA), a practical implementation in the NARS family that emphasizes efficient control and real-time use in applied settings. ONA serves as the runtime system for the benchmark: once a Narsese program is constructed, it is executed by ONA, and the resulting output is used to determine the final label. In this way, ONA is not only the target reasoning engine used at evaluation time, but also the system used to validate whether the compiled symbolic program behaves as intended.

Once a Narsese program is constructed, it is executed in ONA. In the current pipeline, the program body is submitted first, followed by a fixed number of inference cycles before the query is posed. The implementation uses 20 cycles before querying the engine. Runtime output is then inspected to recover the answer signal.

To convert ONA's output into the three-label evaluation space, we map the returned frequency to a categorical label using fixed thresholds. If the recovered frequency is at least 0.50, the answer is mapped to \textsc{True}; if it is at most 0.05, it is mapped to \textsc{False}; otherwise it is mapped to \textsc{Uncertain}. If ONA returns no answer, the instance is also treated as \textsc{Uncertain}. This step converts the reasoner's graded output into the label format used throughout the benchmark.

An example is retained in the released benchmark only if the executed label matches the gold label inherited from the source reasoning instance. This filtering step is important because it ensures that each retained example is not only syntactically well formed, but also behaviorally validated under the same execution protocol that will later be used for evaluation.

This gives the dataset an unusual property for a language reasoning benchmark: every retained symbolic target has been checked by the runtime system that will later be used for evaluation. As a result, the benchmark evaluates not only whether a symbolic program can be generated, but also whether that program executes correctly in the intended reasoning engine.

\section{A Trainable Neuro-Symbolic Extension}

Although the benchmark can already be used as a pure evaluation resource, the same setup also supports learning. A model may take a natural-language context and query, generate FOL or Narsese, and then be trained not only on symbolic similarity to a reference target, but also on whether its output executes correctly.

Let $z^{*}_{\mathrm{FOL}}$ denote the reference FOL target, and let $z_{\mathrm{FOL}}^{(1)}, \dots, z_{\mathrm{FOL}}^{(k)}$ denote sampled candidate generations. We define a standard token-level loss $\mathcal{L}_{\mathrm{text}}$ against the reference sequence and an execution-based loss $\mathcal{L}_{\mathrm{exec}}$ that rewards candidates whose compiled-and-executed outputs yield the correct label. The combined objective is
\[
\mathcal{L} = \mathcal{L}_{\mathrm{text}} + \lambda \mathcal{L}_{\mathrm{exec}}.
\]

The motivation is straightforward. Two symbolic strings may look different on the surface while still being equivalent at execution time, and a string that appears plausible may still fail to execute or may produce the wrong reasoning outcome. Execution therefore provides a task-level semantic signal that complements ordinary sequence supervision. In this sense, the benchmark can support both evaluation and training: it measures whether a model can generate symbolic structure that is not only well formed, but also executable and semantically correct.

The same dataset can also support a simpler but practically useful supervised setting in which the model predicts the final reasoning label directly from natural language. As an initial proof of concept, we trained and released a compact LoRA adapter on top of \texttt{microsoft/phi-2}, using \textit{NARS-Reasoning-v0.1} as the supervision source. In this setup, the model reads a natural-language context and claim and predicts one of three labels: \texttt{A} (\textsc{True}), \texttt{B} (\textsc{False}), or \texttt{C} (\textsc{Uncertain}). The released adapter uses PEFT-based LoRA on a 4-bit NF4 quantized Phi-2 base model, with prompt masking and answer-only supervision, and was published as \texttt{MinaGabriel/phi2-2.7b-lora-nars-adapter} \cite{gabriel2025phi2adapter,phi2_modelcard}.

This released adapter should be viewed as a first-stage neuro-symbolic baseline rather than the final form of the proposed pipeline. It demonstrates that the benchmark can already support lightweight supervised training with a small model, and that the natural-language inputs in the dataset contain enough structure to learn nontrivial three-label reasoning behavior. At the same time, the adapter is still a direct classification model: it predicts labels rather than generating full executable symbolic programs.

The broader research direction is therefore two-stage. In the first stage, a compact model can be trained to solve the task as three-label reasoning classification, which provides a practical baseline and a strong sanity check on dataset quality. In the second stage, the same benchmark can be used to move beyond label prediction and toward symbolic generation, where the model is trained to produce FOL or Narsese explicitly and is rewarded based on runtime correctness. Under that stronger setting, the training target is no longer only the final answer, but the executable reasoning structure itself.

\section{Evaluation Protocol}

The primary evaluation setting is strict. At test time, the system receives only the natural-language context and query. It must generate a syntactically valid and semantically usable symbolic program, execute it in a NARS engine, and return one of \{\textsc{True}, \textsc{False}, \textsc{Uncertain}\}. Producing an intermediate FOL representation is allowed for interpretability, but it is not required.

We recommend reporting overall accuracy, accuracy by difficulty level, macro-averaged F1 over the three labels, confusion matrices, and execution success rate. The last metric is especially important. A system that reaches the correct answer by shortcut guessing but rarely emits executable symbolic programs should not be considered equivalent to a system whose outputs support actual reasoning.

The benchmark supports several baseline families. The first consists of direct-answer LLMs that read the natural-language problem and predict \textsc{True}, \textsc{False}, or \textsc{Uncertain} without symbolic execution. The second consists of NL$\rightarrow$FOL systems, which generate FOL and are evaluated either by symbolic equivalence or by downstream compilation. The third consists of NL$\rightarrow$Narsese systems, which directly emit executable Narsese programs that are run in ONA.

In addition to these baseline families, the benchmark already supports a trained lightweight baseline built directly on the released dataset. Our Phi-2 LoRA adapter provides a concrete supervised benchmark point: unlike zero-shot prompting baselines, it has been explicitly adapted to the reasoning structure and three-label output space of \textit{NARS-Reasoning-v0.1}. This makes it useful as a bridge between ordinary language-model classification and the stronger executable evaluation setting proposed in this paper.

\subsection{Initial Supervised Baseline}

As an initial supervised experiment, we fine-tuned a LoRA adapter on top of \texttt{microsoft/phi-2} using \textit{NARS-Reasoning-v0.1}. The model was prompted with a natural-language context and claim and trained to output one of the three letters \texttt{A}/\texttt{B}/\texttt{C}, corresponding to \textsc{True}/\textsc{False}/\textsc{Uncertain}. This design intentionally keeps the first baseline simple: it tests whether a compact instruction-following model can internalize the reasoning patterns in the dataset before full symbolic generation is introduced.

The resulting adapter showed encouraging behavior in pilot experiments. In particular, it demonstrated that a compact 2.7B base model can be adapted to the dataset-specific reasoning format and can produce stable three-label predictions under deterministic decoding. This does not yet establish executable symbolic reasoning, but it does provide evidence that the benchmark is learnable in a supervised setting and that it supports practical small-model experiments.

From the perspective of this paper, the main importance of this result is methodological. The benchmark is not only a filtered evaluation set paired with executable symbolic programs; it is also a usable supervised training resource. This matters because it opens two complementary research paths: one path studies direct reasoning classification from natural language, and the other studies full symbolic generation followed by runtime execution. The former provides a compact, reproducible baseline; the latter is the stronger long-term objective of the benchmark.

When quantitative results are reported, we recommend presenting the Phi-2 LoRA adapter as a direct-answer supervised baseline and comparing it against zero-shot instruction models of similar and larger scale. This comparison helps separate gains due to dataset-specific adaptation from gains due only to model size or general pretraining.

\subsection{Why Executable Evaluation Matters}

A major weakness of ordinary reasoning benchmarks is that a correct final answer can hide an incorrect internal representation. By contrast, executable evaluation forces the system to preserve structural information that matters. If the symbolic program is malformed, over-strengthened, under-specified, or semantically misaligned, execution will expose that failure.

In that sense, the benchmark tests something stronger than answer selection: it tests whether a model can translate language into a representation that another reasoning system can actually use.

\section{Discussion}

This work should be understood as a step toward tighter integration between language models and reasoning engines, not as a claim that unrestricted natural language can already be mapped perfectly into Narsese. The compiler supports a benchmark-specific subset of FOL patterns, and the dataset is filtered to retain instances that execute correctly in ONA. This is a strength for benchmark reliability, but also a limitation for generality.

Several trade-offs follow from this design. First, some compiler mappings intentionally strengthen or decompose expressions in order to fit executable Narsese patterns. This is acceptable for a controlled benchmark, but it means the system is not yet a fully general semantics-preserving translator. Second, the execution protocol depends on ONA's runtime behavior, including the chosen cycle budget and the thresholds used to map graded outputs into three labels. Different budgets or thresholds may change the observed distribution of \textsc{Uncertain} outcomes. Third, while the benchmark is derived from a strong FOL reasoning source, it remains synthetic in origin through the ProverGen framework. As with other synthetic reasoning datasets, strong performance here does not automatically guarantee robust performance on unconstrained real-world language.

At the same time, the existence of a released trained adapter changes the practical status of the benchmark. The project is no longer only a benchmark proposal plus a compiler; it also includes an initial trained model built directly on the dataset. This strengthens the paper because it shows that the resource is already usable in practice. The Phi-2 LoRA baseline demonstrates that the dataset can support compact supervised learning today, while the execution-based pipeline defines a stronger next step in which models generate explicit symbolic forms that are validated at runtime.

Even with its current limits, the benchmark fills a useful niche. It connects natural-language reasoning, symbolic compilation, executable validation, and supervised adaptation in a single pipeline. This makes it relevant to AGI-oriented research, where the goal is not merely fluent language generation, but language-conditioned reasoning that can interact with a persistent symbolic substrate.

\section{Conclusion}

We presented \textit{NARS-Reasoning-v0.1}, a benchmark of 1{,}000 natural-language reasoning instances paired with executable Narsese programs, together with a deterministic compiler from FOL to Narsese and an execution-based validation procedure in ONA. We also described a trainable neuro-symbolic extension in which language models are encouraged to generate executable reasoning structure rather than only verbal answers.

In addition, we showed that the benchmark already supports supervised adaptation in practice by training and publishing a Phi-2 LoRA adapter for three-label reasoning classification on the dataset. This initial model does not yet generate full executable symbolic programs, but it demonstrates that the benchmark is not only theoretically motivated; it is also directly usable for model development and baseline construction.

\bibliographystyle{splncs04}
\bibliography{references}

\end{document}